\documentstyle[leqno,pubsmacs,pubsbib]{pubsarticle}
\newcommand{\imp}{\mbox{$\rightarrow$}}
\newcommand{\lh}{\langle}
\newcommand{\rh}{\rangle}
\newcommand{\bs}{\backslash}
\def\SUBCAT{\small\rm SUBCAT}
\def\BMW{\small\rm BMW}
\def\HBU{\small\rm HBU}
\def\NP{\small\rm NP}
\def\VP{\small\rm VP}
\def\N{\small\rm N}
\def\S{\small\rm S}
\def\X{\small\rm X}
\def\Y{\small\rm Y}

\title{A compositional treatment of polysemous arguments in Categorial
Grammar}
\author{Anne-Marie Mineur\thanks{
	Computational Linguistics, University of the Saarland,
	Postfach 1150, 66041 Saarbr\"ucken, Germany}\ \
\and Paul Buitelaar\thanks{
	Computer Science, Brandeis University, Waltham, MA 02254, USA}}
\date{\small\tt (mineur@coli.uni-sb.de ; paulb@zag.cs.brandeis.edu)}

\makeindex

\begin{document}
\maketitle

% This in order to make the title fit as a header.
\makeatletter
	\def\rightmark{A compositional treatment of polysemous arguments in CG}
	\if@twoside\def\leftmark{A compositional treatment of polysemous arguments in
CG}\fi
\makeatother

%------------------------------Abstract-----------------------------------

\begin{abstract}
We discuss an extension of the standard logical rules (functional
application and abstraction) in Categorial Grammar (CG), in order to deal
with some specific cases of polysemy. We borrow from Generative Lexicon
theory which proposes the mechanism of {\em coercion}, next to a rich
nominal lexical semantic structure called {\em qualia structure}.

In a previous paper we introduced coercion into the framework of {\em
sign-based} Categorial Grammar and investigated its impact on
traditional Fregean compositionality. In this paper we will elaborate on
this idea, mostly working towards the introduction of a new semantic
dimension. Where in current versions of sign-based Categorial Grammar
only two representations are derived: a prosodic one (form) and a
logical one (modelling), here we introduce also a more detaled
representation of the lexical semantics. This extra knowledge will serve
to account for linguistic phenomena like {\em metonymy\/}.
\end{abstract}

%-------------------------1. Introduction---------------------------------

\section{Introduction}

Categorial Grammar\index{Categorial Grammar} is a
compositional\index{compositional} theory. The representation of the
meaning of a whole is a function of the representation of the meaning of
its parts. Since Categorial Grammar has only been interested in
modelling predicate-argument structure, scope and the like, a simple
representation of the nominal lexical semantics was sufficient in most
cases\footnote{
	Although some of the work on {\em meaning postulates} developed
	a more elaborate lexical semantics (\cite{dowt:word79}).}.
As observed by several researchers however, such a basic Fregean concern
with composition is not enough for a cognitive or computational system
(\cite{pust:gene95}; \cite{sowa:logi92}).  Frequently there is
additional information in between the composing parts, which is inferred
by heuristic reasoning, e.g., by {\em metonymy} as in
(\ref{metonymy}),  which has to be interpreted as `John beginning to
do something with/to a novel', probably `reading' or `writing' it
(\cite{pust:gene95}).

\enumsentence{\label{metonymy}John began a novel.}

To account for this kind of inferences in Categorial Grammar some
additions have to be made to its present way of representing word
meaning. We borrow from Generative Lexicon theory\index{Generative
Lexicon Theory} (\cite{pust:gene95}) in defining the multiple meanings
of a word by the use of a polymorphic\index{polymorphism} representation.
Instead of giving a traditional enumeration of word senses, we now
relate them to one another into one coherent structure ({\em qualia
structure})\index{Qualia Structure}. This also allows for a generation
of senses in context, that is in a dynamic fashion by means of {\em
coercion}\index{coercion}. We see this in an object-oriented way, where
objects derive their meaning from the procedures that operate on
them. Each procedure addresses a different aspect or quality ({\em
`qual'\/})\index{qual} of an object.

We want to stress that we are not talking about syntactic polymorphism,
like a noun that is ambiguous between nominative and accusative
(\cite{pull:phon86}; \cite{ingr:limi90}; \cite{baye:coor94}).  In this
paper we are interested in how the meaning of one word influences that
of another and thus how semantics is built up by association while still
maintaining compositionality.  In fact what we are saying, is that there
are no such things as uninterpretable, syntactically well-formed
sentences. To our minds the sentence `Colourless green ideas sleep
furiously', that Chomsky (\shcite{chom:synt57}) claimed proved the
autonomy of syntax, only proves that people assign an interpretation to
almost any syntactically well-formed sentence---and that they have a
very flexible imagination.

In short, form (morphology) and structure (syntax) serve meaning, in the
sense that they indicate to the hearer how to interpret the concepts
that are communicated, and in what way those are related. The less
well-formed a sentence, the less clear it is which relation exists
between which objects, and the more complicated to fit
subcategorization\index{subcategorization} and argument. Once it is
clear which argument belongs to which subcategorization, the limits to
coercion seem very flexible.

 From a computational angle however certain limits have to be dealt
with.  Currently it does not seem feasible to compute whatever
interpretation for any given syntactically well-formed sentence. This is
why computational linguistic theories like Categorial Grammar seem
rather to ignore semantic anomalies as they occur in sentences like
(\ref{metonymy}). In this paper however we try to deal with them in some
computationally acceptable way, while at the same time acknowledging the
fact that not every possible interpretation can or even should be
computed. If one assumes a semantic network for a given language in
which words are ultimately (through recursive links) related to every
other word in the language, it must be obvious that not every
interpretation to be obtained from this network should be computed.
Therefore we take the simple but effective heuristic of computing only
one level deep. Only the semantic elements to be found in the immediate
qualia structure of a word are to be considered for interpretation,
which takes the form of a selection of the right interpretation(s) from
this lexical semantic structure. The compositional process selects the
appropriate semantic element(s) ({\em sorts})\index{sorts} out of the
qualia structure, which is represented as a {\em set}. Such an
approach reinterprets Pustejovsky's {\em coercion} as a form of type
{\em selection} rather than type {\em shifting}, which makes it also a
monotonic operation\footnote{
	The observation that in such an approach coercion becomes a
	monotonic operation is due to Hans Uszkoreit (personal
	communication).
}.

The reason for choosing Categorial Grammar as the grammar formalism is
rather arbitrary, although not entirely. Both Categorial Grammar and
Generative Lexicon theory are descendants from Montague Grammar, and
share a number of assumptions (lexicalism,\index{lexicalism}
compositionality). The multi-level approach that is pursued nowadays
in sign-based Categorial Grammar facilitates adding an extra
autonomous level, and constitutes an interesting extension to
Categorial Grammar.

%------------------------2. Compositionality------------------------------

\section{Compositionality}

\subsection{Functors and Arguments}

In Categorial Grammar one defines a category\index{category} in terms of
its domain and its yield. An intransitive verb requires a noun phrase to
yield a sentence; a transitive verb requires a noun phrase to yield an
intransitive verb; an article requires a noun to yield a noun phrase.
Even complex arguments are possible---take for instance a modifier to a
{\VP} (i.e., an {\NP}$\bs${\S}): it requires an {\NP}$\bs${\S} to yield
an {\NP}$\bs${\S}.

\enumsentence{\label{categories}
\begin{tabular}[t]{ll}
	Intransitive verb: & {\NP}$\bs${\S}\\
	Article:	   & {\NP}/{\N}\\
	Transitive verb:   & ({\NP}$\bs${\S})/{\NP}\\
	{\VP}modifier:     & ({\NP}$\bs${\S})/({\NP}$\bs${\S})
\end{tabular}
}
Directionality\index{directionality} is indicated by the direction of
the slash; an {\X}/{\Y} ({\em `{\X} over {\Y}'}\/) seeks a {\Y} to its
right to yield an {\X}, whereas a {\Y}$\bs${\X} ({\em `{\Y} under
{\X}'\/}) seeks a {\Y} to its left to yield an {\X}.

The basic categories we use are {\N} (for noun), {\NP} (for noun phrase)
and {\S} (for sentence). All basic categories are categories. Complex
categories are of the form {\X}/{\Y} or {\X}$\bs${\Y}, where {\X} and
{\Y} must also be categories.

This categorial approach has two immediate effects, first that all the
combinatorial information is transferred to the lexicon, so that every
entry has all the information it needs ({\em lexicalization\/}),
secondly that the set of rules is reduced to two rules, that do nothing
more than describe the behavior of the connectives\index{connectives}
{\bf /} (`slash') and~{\bf $\bs$}~(`backslash'):

\enumsentence{\label{simple_rules}\shortexnt{1}
	{{\X}/{\Y}, {\Y} \imp\ {\X}}
	{{\Y}, {\Y}$\bs${\X} \imp\ {\X}}}

\subsection{Logical Representation}

It follows naturally to extend this compositional approach of
constituent structure with a logical representation\index{logical
representation}. Since in Categorial Grammar the choice to call
something a functor\index{functor} is usually based both on content and
on form, we find the same division as is found in the categories to be
reflected in the logical representation. The result of applying a
functor to an argument in terms of a logical form is the {\em logical
representation\/} of the functor applied to the {\em logical
representation\/} of the argument.

\enumsentence{\label{labeled_categories}\shortexnt{1}
	{$\lh$ ({\X}/{\Y}) : {\em functor} $\rh$,
	$\lh$ {\Y} : {\em argument}$\rh$ \imp\
	$\lh$ {\X} : {\em functor(argument)}$\rh$}
	{$\lh$ {\Y} : {\em argument} $\rh$,
	$\lh$ ({\Y}$\bs${\X}) : {\em functor} $\rh$ \imp\
	$\lh$ {\X} : {\em functor(argument)}$\rh$}
}
As we use these rules recursively, and as each of the lexical entries
carries its own bit of information, the logical representation of the
whole is composed {\em exclusively\/} of the logical representation of
the subparts. If we follow the traditional Montagovian way of
representing meaning, then the lexical entries for `explained, `a
speaker' and `an example' would be:

\enumsentence{\label{assignments}
\begin{tabular}[t]{llll}
{\bf explained} & $\leadsto$ & ({\NP}$\bs${\S})/{\NP} & : $\lambda
		qp.explain(p,q)$\\
{\bf an example} & $\leadsto$ & {\NP} & : $\lambda R.\exists
		y.example(y) \wedge R(y)$\\
{\bf a speaker} & $\leadsto$ &  {\S}/({\NP}$\bs${\S}) & :
		$\lambda P.\exists x.speaker(x) \wedge P(x)$
\end{tabular}
}
Then the representation of (\ref{1a}) would be derived
straightforwardly as (\ref{1b}). (Albeit with the intervention of some
simple meaning postulates to account for the equivalence of a term and
its $\eta$-normal form, see (\cite{mont:form74}):

\eenumsentence{
\item\label{1a}
	 A speaker explained an example.
\item\label{1b}
	$\exists x.speaker(x) \wedge \exists y.example(y) \wedge explain(x,y)$
}
This lambda-term\index{lambda-term} can on the one hand be seen as the
output of the derivation\index{derivation} procedure (which functors
applied to which arguments) and on the other hand as the input for the
interpretation function (the interpretation in the model).

However, there are some shortcomings to this approach. First, (\ref{1b})
can hardly be considered an exhaustive representation of (\ref{1a}). It
does not show that a `speaker' is a person; nor does it represent the
information that `to explain' means that you exemplify something to an
audience; nor that an `example' is something that is chosen because of
its typical properties or behaviour. Secondly, nothing prevents
sentences like (\ref{2a}) through (\ref{2c}):

\eenumsentence{
\item\label{2a} An example explained an example
\item\label{2b} A speaker explained a speaker
\item\label{2c} An example explained a speaker
}
These sentences have some semantic mismatch in them, but---and this is
an important observation---we still can construct some meaning, albeit
with increasing difficulty. That is, if we use all the knowledge that we
have available on each of these words. As was mentioned before, meaning
is more than an addition of the single atomic senses.

The idea of a semantic mismatch is mirrored quite well in the way some
semanticists (e.g., \cite{poll:info87}) formulate it: `explain' has not
just syntactic expectations considering its arguments, but also expects
their meaning to meet certain {\em selection restrictions}, see
(\ref{sem-sel}). The first {\sc subcat} {\NP} has to be fulfill the role
of the explainer, the second that of the explained.

\enumsentence{\label{sem-sel}
	explain $\rightarrow$
	[ {\SUBCAT} : $\langle$
                      {\NP}$_{\fbox{\footnotesize 1} \ explainer}$,
                      {\NP}$_{\fbox{\footnotesize 2} \ explained}\rangle$]
}
If we look at the representation in~(\ref{1b}) none of this shows.
All (\ref{1b}) does is indicate what the predicate
is and what the arguments, and what relations there are between them.

%----------------------3. Lexical Semantics-------------------------------

\section{Lexical Semantics}

\subsection{Towards a Lexical Semantics}

If we want to talk about natural language not only in terms of
predicate-argument schemata but also in terms of associations and the
bridging of semantic gaps, then we have to consider {\em all\/} the
semantic relations that exist between the words in the lexicon. We have
to take into account not only verbs and their arguments, but also the
lexical semantics of other categories, more in particular of nouns.

We claim that it is virtually impossible for any word to have only one
strict interpretation. By their very nature words adapt to their
context.  The hearer as well as the speaker use their imagination and
select the one aspect of the concept (relating to a particular word)
that fits best to the requirements of the situation (see also
(\cite{bart:cons87})). In short: polysemy is everywhere.

In the following sections we will discuss the systematic nature of this
semantic polymorphism, as well as the way in which it interacts with
\mbox{context}.

\subsection{Polysemy and Complementary Versus Contrastive Senses}
\label{3.2}

As observed by Weinreich (\shcite{wein:webs64}), distinctive
interpretations or {\em senses} of words are either of a {\em
contrastive}\index{contrastive} or a {\em
complementary}\index{complementary} nature. Contrastive senses (or {\em
homonyms})\index{homonyms} are unrelated to each other, see for
instance the different meanings of the word `bank' in (\ref{3a}) and
(\ref{3b}):

\eenumsentence{
\item\label{3a} We walked along the bank of the Charles river
\item\label{3b} Did he have an account at the {\HBU} bank?
}
Complementary senses on the other hand do not contrast each other, but
seem to be related in a systematic way. For instance a {\em brand name}
like `{\BMW}' has at least the following complementary senses\footnote{
	Note that (\ref{4b}) through (\ref{4d}) refer to the company,
	hence `BMW' is used as a proper name, where (\ref{4a}) and
	(\ref{4e}) refer to the car BMW, and hence `BMW' is used
	as a common noun.
}:

\eenumsentence{
\item\label{4b}\shortexnt{1}
	{{\em the company that produces it}:}
	{{\BMW} stocks gained two points yesterday.}
\item\label{4c}\shortexnt{1}
	{{\em the company building}:}
	{{\BMW} takes up half this block.}
\item\label{4d}\shortexnt{1}
	{{\em a spokesperson with the company}:}
	{{\BMW} announced a new model last week.}
\item\label{4a}\shortexnt{1}
	{{\em the product}:}
	{This year around 10,000 {\BMW}s will be sold.}
\item\label{4e}\shortexnt{1}
	{{\em the design or production process}:}
	{They started a new {\BMW} last year.}
}
The systematic relation between complementary senses has to some extent
been investigated in the literature (\cite{wein:webs64},
\cite{apre:regu73}, \cite{nunb:nonn79}, \cite{bier:sema82}); most
recently by Pustejovsky (\shcite{pust:gene95}) who termed it {\em
logical polysemy}. Some other examples are: \\[2ex]
{\sc Institutions}
In the case of {\em institutions} we quote the well known `school'
example (\cite{bier:sema82}):

\eenumsentence{
\item\label{5a}\shortexnt{1}
	{{\em as a group of people}:}
	{The school went for an outing}
\item\label{5b}\shortexnt{1}
	{{\em as a learning process}:}
	{School starts at 8.30}
\item\label{5c}\shortexnt{1}
	{{\em as an institution}:}
	{The school was founded in 1910}
\item\label{5d}\shortexnt{1}
	{{\em as a building}:}
	{The school has a new roof}
}
{\sc Artifact-Event}
In the following sentences we see a {\em metonymic}\index{metonymy}
relation between an {\em artifact} (`novel') and an {\em event}
(`reading') that typically involves that {\em artifact}.

\eenumsentence{
\item\label{6a} John began the novel
\item\label{6b} John began reading the novel
}
This is a general pattern, which is productive for all {\em
artifact\/} denoting words. Another example is a `model' and the
{\em event\/} of `producing' it.

\eenumsentence{
\item\label{7a} The president of {\BMW} announced a new model
\item\label{7c} The president of {\BMW} announced that they will
                      produce a new model
}
{\sc Figure-Ground} \mbox{} Pustejovsky and Anick (\shcite{pust:sema88})
give several examples of so-called {\em
figure-ground}\index{figure-ground nominals} nominals that are either to
be interpreted as the physical object themselves (figure) or the open
space they leave behind upon removal (ground).

\eenumsentence{
\item\label{9a}\shortexnt{1}
	{{\em figure}:}
	{John painted the door blue}
\item\label{9b}\shortexnt{1}
	{{\em ground}:}
	{John walked through the door quickly}
}
Interestingly, we can also construct the following perfectly acceptable
example where {\em figure\/} and {\em ground\/} interpretations occur at
the same time:

\eenumsentence{\addtocounter{enums}{-1}
\item[c.]\label{9c}
	Go through the red door on your right.
}
That this can not always be done is shown by the following two
examples with {\sl book}, which is both of sort {\em information} and
{\em physical object}---apparently {\sl to buy} doesn't trigger the
{\em physical object}-aspect of {\sl book}, or not as much as {\sl to
throw} does.

\eenumsentence{
\item\label{9d} \ \ \ \ \  I love the book you bought me.
\item\label{9e} ?? I love the book you threw at me.
}
{\sc Animal-Food} {\em Grinding}\index{grinding} (see
e.g., (\cite{cope:lexi92})) is the systematic relation between an {\em
animal} and the {\em food} those animals produce after being killed.

\eenumsentence{
\item\label{10a}\shortexnt{1}
	{{\em the animal}:}
	{You won't find {\em badgers} living around here.}
\item\label{10b}\shortexnt{1}
	{{\em the food it produces}:}
	{{\em Badger} is a delicacy in China.}}

\subsection{Lexical Semantics and Context}

Natural language is at the same time the most powerful and the most
limited knowledge representation language. Most powerful, because in no
artificial language can we express what we mean so vividly and full of
interpretations. Most limited, because the number of possible
interpretations that are somehow conveyed in natural language is just
too big to yield any precision. This would mean that one is never able
to point out what exactly the meaning of an expression is. However,
humans are able to communicate in a sensible way most of the time. This
leads us to believe that in fact there are two levels of semantic
reasoning, one operating on information that can be obtained from the
lexicon, and one that involves real-world knowledge\index{real-world
knowledge}. We illustrate this distinction with the following example:

\enumsentence{\label{bi-qua}
	The newspaper that fell off the table, fired its chief editor.
}
Technically speaking sentence (\ref{bi-qua}) could have a meaning: one
might imagine a talking, acting, living newspaper that can both fall off
the table and fire people\footnote{
	The proposition that we do not allow any word to have an
	ambiguous meaning, or, in our case to use multiple aspects of
	its meaning at a time, is motivated extensively by Bayer
	(\shcite{baye:coor94}). There is reason to believe
	that this proposition does not hold for all situations, take
	for instance example~(\ref{9c}), as well as the following
	sentence: "The school that starts at 9 am, burnt down this
	morning", which uses two aspects of the word's meaning that
	apparently do unify.},
but given our knowledge about the physical world, it is not a very
likely one. In other words: in our theory {\sl newspaper} will be
coerced to {\em human} and then some interpretation can be assigned to
the sentence, but unless we use it in an Alice-in-Wonderland setting,
real-world knowledge will rule it out.

In this paper we will only deal with the compositionally construable
aspects of systematic polysemy without looking at any specific context
dependent interpretation.

%---------------------4. Lexical Semantic Structure--------------------

\section{Lexical Semantic Structure}

\subsection{Four Levels of Interpretation}

We adopt Pustejovsky's (\shcite{pust:gene95}) model of lexical semantic
structure, which  introduces four interrelated levels of interpretation:

\paragraph{Argument Structure}\index{Argument Structure} Specification
of number and type of  logical arguments, and how they are realized
syntactically\footnote{
	Notice that there is a difference between Pustejovsky's
	(\shcite{pust:gene95}) notion of Argument Structure, where three
	types of arguments are distinguished (logical, default and
	shadow arguments) and the CG notion of predicate-argument
	structure, which only accounts for those arguments that are
	actually realized---logical arguments, in Pustejovsky's terms.
}.

\paragraph{Event Structure}\index{Event Structure} Definition of what
sort of event a lexical item represents. Sorts include {\sc state}, {\sc
process} and {\sc transition}. Events may have sub-eventual structure.

\paragraph{Qualia Structure} Modes of explanation, composed of {\sc
formal}, {\sc constitutive}, {\sc telic} and {\sc agentive} roles.  In
more common AI terms these roles correspond to the {\sc is-a}, {\sc
part-of/has-a}, {\sc purpose} and {\sc cause} relations respectively.

\paragraph{Lexical Inheritance Structure}\index{Lexical Inheritance
Structure} Identification of how a lexical object is hierarchically
related to other objects in a lattice that constitutes the global
organization of the lexicon.

\subsection{Qualia Structure}

In \cite{pust:gene95} qualia structure is seen as a polymorphic
representation with the different qualia roles expressing different
aspects of the lexical semantic object that it represents. An example
is a qualia structure for `{\BMW}' given in figure~(\ref{bmw-qualia}).
The values of the qualia roles are meant to be defaults (the ``\dots '' are
to be filled in depending on context). They reflect typical knowledge
related to the concept `{\BMW}'.

\enumsentence{\label{bmw-qualia}
   \avm{{\sc formal}        & :  $company(bmw)$ \\
	{\sc constitutive}  & :  $\exists y.spokesperson\_of(y,bmw), \dots$ \\
	{\sc agentive}      & :  $\exists e.\exists x.human(x) \wedge
				    establish(e,x,bmw), \dots$ \\
	{\sc telic}         & :  $\exists e.\exists z.produce(e,bmw,z), \dots$}
}
This structure reads as follows: If we take {\BMW} as a company ({\sc
formal}), then depending on what qualia role is highlighted, it can
also be seen as the group of people who constitute it, one of whom is
the spokesperson ({\sc constitutive}); it entails the process of being
established ({\sc agentive}); and it entails its default purpose,
which is the production of cars ({\sc telic}). In other words, in this
case `{\BMW}' is polymorphic between the predicates `company',
`spokesperson\_of', `establish' and `produce'.

\subsection{Coercion}

If qualia structure is to be seen as a polymorphic representation, then
{\em coercion} is the generation process that produces each individual
sense that is represented in the qualia structure. This notion of {\em
coercion} is similar to the original notion of the same name in the
context of object-oriented and functional programming languages
(\cite{card:unde85}).

Essentially, the original notion of \mbox{coercion} is restricted to
changing the type of an object if a particular function that takes the
object as input requires it. From the standpoint of natural language
semantics this can be seen as some form of interpretation. The context
of the object forces, {\em coerces\/} it to be interpreted
differently, to take on another denotation\index{denotation}.

In natural language interpretation this happens all the time, but
unlike \cite{pust:gene95}, we prefer to see this as a selection out of
a set of possible interpretations rather than as some form of meaning
shift. In the sentences on {\BMW} above we saw an example of such a
set of interpretations to choose from: a company, a car, a building, a
collection of people and a process.

In the following section we will try to formalize this in a Categorial
Grammar framework.

%-------------5. A Categorial Treatment of Polysemous Arguments--------

\section{A Categorial Treatment of Polysemous Arguments}

\subsection{Motivation}

The rules for functional application in {\em traditional\/} Categorial
Grammar (\cite{barh:quas53}), when labeled with a logical
representation, as given in (\ref{labeled_categories}) are repeated here:

\enumsentence{\label{lab-func-app}\shortexnt{1}
	{$\lh$ {\X}/{\Y} : {\em f} $\rh$, $\lh$ {\Y} : {\em a} $\rh$
		 \imp\ $\lh$ {\X} : {\em f(a)} $\rh$}
	{$\lh$ {\Y} : {\em a} $\rh$, $\lh$ {\Y}$\bs${\X} : {\em f} $\rh$
		\imp\ $\lh$ {\X} : {\em f(a)} $\rh$ }
}
That is, a type {\X} can be derived from the concatenation of a functor
{\X}/{\Y} with an argument {\Y} on its right, or a functor {\Y}$\bs${\X}
with an argument {\Y} on its left. In terms of its logical
representation the functor is applied to the argument.

But now we are proposing to insert lexical semantic information (qualia
structure) into these combinatorial rules. We want to do it in such a
way that it does not influence the derivability\index{derivability} of a
sentence, nor the compositionality of the logical representation.
Whether we want the lexical semantic information to be reflected in the
logical representation is another matter. We insist on
maintaining compositionality. However, since our representation of the
composing parts (i.e., the lexical objects) will be more elaborate,
this may effect the representation of the whole (i.e., the sentence),
and it may in fact influence its denotation.

\subsection{The Categorial Sign}
\label{CatSign}

The nucleus of the categorial sign is the category, which describes
the combinatorial behaviour. To it may be attached a number of labels,
which represent different dimensions (e.g., denotational semantics,
prosody) (Gabbay \shcite{gabb:labe91:}).  The rules that describe the
behaviour of the categories will also specify the combinatorial
behaviour of the representation of the other dimensions. Properties
that apply to the category need not apply to the labels: the way a
sequence is proven to be derivable does not matter for the prosodics
of the utterance, and for the representation of the denotation of an
utterance it does not matter in which order the arguments were
found. In other words: bracketing can be omitted in the prosodics
dimension, order is irrelevant for predicate-argument structure.

In sign-based Categorial Grammar (\cite{moor:labe92}, \cite{morr:type94}),
every lexical entry is represented by a sign that contains two
dimensions: prosody\index{prosody} and denotational semantics. The sign,
the basic element of the framework, looks as follows:
\enumsentence{\label{sign-classic}
$\lh$ C : ($\lambda$,P) $\rh$
\begin{tabbing}
where \= \kill
where C = combinatorics;\\
\>    $\lambda$ = denotational semantics;\\
\>    P = prosody
\end{tabbing}
}
The combinatorics\index{combinatorics} is the category as it is
inductively built up from the basic categories {\NP}, {\N} and {\S}. The
basic categories will carry with them those features that can influence
the combinatorial properties of the word, whether they are semantically
relevant or not. Number, person, gender and case are typically
combinatorially relevant properties. The denotational semantics is in
fact the predicate-argument structure of the sentence; the constants
that are used are only mnemonic identifiers, as we know them from the
Montague-literature (\cite{mont:form74}). The prosodic dimension
represents word order, and could be extended to represent constituent
structure and intonation.

Denotational semantics is a standard label in the existing work on
sign-based Categorial Grammar. For the sake of simplicity we will omit
prosody. The categorial sign as we use it then becomes:
\enumsentence{\label{sign-new}
$\lh$ C : ($\lambda$,$\cal{Q}$) $\rh$
\begin{tabbing}
where \= \kill
where C = combinatorics;\\
\>    $\lambda$ = denotational semantics;\\
\>    $\cal{Q}$ = qualia structure
\end{tabbing}
}

The introduction of a qualia structure is new. The qualia structure of
a phrase will consist of the information of its composing parts; it
grows as more lexical objects are added. The interpretation of the
lexical objects however gets more precise: context disambiguates.

\subsection{The Representation of Qualia Structure}

In (\cite{buit:coer94}) we explored the consequences of including the
qualia structure in the {\em combinatorics}, together with all the other
features that described the category's lexical properties. In this way,
we gave the qualia structure combinatorial power, and made it possible
for an application to fail because of conflicting qualia structures.
However, this is not consistent with what we stated above, namely that
people will assign an interpretation to any syntactically well-formed
sentence\footnote{
	 In this respect it is interesting to compare
	(\cite{vers:flex94}), who explores metonymy in a many-sorted
	version of Categorial Grammar---using sorts instead of types
	as basic categories. Like in our current approach this system
	ties syntactically well-formedness to a flexible form of
	semantic interpretation.
}.  Chomsky's example shows that its lexical semantics will not make a
sentence ungrammatical, the arguments will always adapt to their
selectional restrictions---this clearly illustrates the
appropriateness of the term {\em coercion\/}.

In our current proposal qualia structure is an extra semantic
dimension, next to the logical form. It serves to license coercions,
where otherwise a semantic mismatch would have occurred between the
argument and the selection restriction of the functor for that
argument.

Qualia structures are represented as sets of {\em sorts\/}\footnote{
	Specifications on a basic type that correspond to subsets of
	the set of entities that correspond to the basic type in the
	model.}
and are indexed by the {\em entity\/} they correspond to in the model.
For instance:

\enumsentence{
$QS_{\mbox{\em bmw}}$: $\{{\bf company, spokesperson\_of, establish,
produce}\}$
}
Note that with this representation of qualia structure we are abstracting
away from the use of qualia role names like {\sc formal}, etc. (see
figure~(\ref{bmw-qualia}). We acknowledge that subdividing the qualia
set by role names adds additional structure to the lexical semantic
representation, but we refrain from exploiting this for now, in order to
keep the representation legible. Not adding role names, we think, does not
damage the fundamental ideas behind a polymorphic lexical semantic
representation.

Sorts are hierarchically organized in a lattice, where sub-sorts are
subsumed by their super-sorts. In the case where a sub-sort is unified
with its super-sort, the result will be the sub-sort. For instance:
The sub-sorts ${\bf read}$ and ${\bf write}$ share the same super-sort
${\bf event}$, and unification with {\bf event} would result in
{\bf read} and {\bf write}, respectively. In general:
\enumsentence{
${\bf x} \leq {\bf y} \Longrightarrow {\bf x} \sqcup {\bf y} = {\bf x}$.
}
Basic categories carry one qualia structure, while complex categories have
lists of qualia structures (one for the functor and one for the
argument\footnote{
	With the embedding of the categories, the qualia structures of
	complex categories will also be embedded---a transitive verb
	(({\NP}$\bs${\S})/{\NP}) will typically have a qualia structure
	of the form
	\enumsentence{\avm{QS$_{f}$:
				\avm{QS$_{f^\prime}$:\{x$_1$ \dots x$_n$\},\\
				     QS$_{a^\prime}$:\{x$_1$ \dots x$_m$\} },\\
			   QS$_{a}$:\{x$_{1}$ \dots x$_{p}$\}}}
}):
\enumsentence{\label{cat-def}
\begin{tabular}{ll}
	basic & $\lh$ X : ($\lambda$, $QS_{a}:\{q{_1} \dots q{_n}\}$) $\rh$\\
	complex & $\lh$ X/Y : ($\lambda$, $[ QS_{f}:\{q{_1} \dots
		   q_{m}\} , QS_{a}:\{q{_1} \dots q{_p}\} ]$) $\rh$
	\mbox{\rm or}\\
	{}      & $\lh$ Y$\backslash$X  : ($\lambda$, $[ QS_{f}:\{q{_1} \dots
		   q_{m}\} , QS_{a}:\{q_{1} \dots q_{p}\} ]$) $\rh$
\end{tabular}
}

We chose to represent qualia structure as a set and not as a conjunction
or disjunction. The use of conjunction would imply incorrectly that all
values need to be present all the time. The use of disjunction would
imply that values can be maintained, as long as one of the other values
is applicable, even if they are not wanted. Clearly this is not correct
also. We want to be able to make a selection, and only use those values
that apply in a given context.

\subsection{The Calculus}

We will not go into detail about all the varieties of functional
application that occur in the literature. For that we refer the reader
to (\cite{moor:labe92}) and (\cite{morr:type94}) on Lambek-style
Categorial Grammar. What we claim with respect to functional application
of a functor to an adjacent argument, may be generalized to any form of
adjacent or non-adjacent argument resolution.

In order to have a complete calculus, we not only need to have rules for
functional application\index{functional application}, but
abstraction\index{abstraction} (or conditionalization) as well.  The use
of abstraction can best be demonstrated by an example: the derivation of
a {\VP}.  A sequence of types that reduces to form a {\VP} will in
categorial terms be said to reduce to a sentence that lacks a noun
phrase on its left ({\NP}$\backslash${\S}). This sequence can be
reformulated in the following way. We can assume that the missing noun
phrase is present as the leftmost type in the sequence and try to prove
that this sequence combines to form a (complete) sentence. Like in any
other equation, we can simply add a category on both sides. This
category is hypothetical, and so are all the labels that are attached to
it. In the denotational semantics dimension this materializes by the
introduction of a lambda operator.

To represent the relation between the original sequence and the
reformulated one in a rule, we need a notation that can show the
derivability between two sequences. We also need such a notation to
generalize the simple rule for application---we want to be able to
capture any sequence of categories that resolves to the argument. Only
in the final stage, when the categories are reduced to basic categories,
will there be a check for identity between the given category and the
required one---the axiom.  The set of rules for the basic fragment are
based on Moortgat (\shcite{moor:cate88:}) and look as follows.

\subsection{The Original Rules}

Application is given in (\ref{simple_rules_appl}), where a functor {\X}/{\Y}
(or {\Y}$\bs${\X}, respectively), preceded by left context U and followed by
right context V, takes some sequence of categories T on its right (or
on its left, respectively) as its argument if T reduces to {\Y}, and the
resulting category {\X} with U and V reduces to the same Z.
\enumsentence{\label{simple_rules_appl}
\begin{tabular}[t]{llll}
[L/]
&
\begin{minipage}[t]{39mm}
	U, $\lh$ {\X}/{\Y}:{\em f} $\rh$, T, V \imp\ Z if \\
	T \imp\ $\lh${\Y}:{\em a} $\rh$ and\\
	U, $\lh$ {\X}:{\em f(a)} $\rh$, V \imp\ Z
\end{minipage}
&
[L$\bs$]
&
\begin{minipage}[t]{39mm}
	U, T, $\lh$ {\Y}$\bs${\X}:{\em f} $\rh$, V \imp\ Z if \\
	T \imp\ $\lh$ {\Y}:{\em a} $\rh$ and\\
	U, $\lh$ {\X}:{\em f(a)} $\rh$, V \imp\ Z
\end{minipage}
\end{tabular}
}
Abstraction is given in (\ref{simple_rules_abstr}), where a sequence of
categories T reduces to some functor {\X}/{\Y} (or {\Y}$\bs${\X},
respectively) if T followed by {\Y} (or preceded by it, respectively)
reduces to {\X}.

\enumsentence{\label{simple_rules_abstr}
\begin{tabular}[t]{llll}
[R/]
&
\begin{minipage}[t]{4cm}
	T \imp\ $\lh$ {\X}/{\Y}:{\em f} $\rh$ if \\
	T, $\lh$ {\Y}:{\em a} $\rh$ \imp\ $\lh$ {\X}:{\em $\lambda$a.f} $\rh$
\end{minipage}
&
$[$R$\bs]$
&
\begin{minipage}[t]{4cm}
	T \imp\ $\lh$ {\Y}$\bs${\X}:{\em f} $\rh$ if \\
	 $\lh$ {\Y}:{\em a} $\rh$, T \imp\ $\lh$ {\X}:{\em $\lambda$a.f} $\rh$
\end{minipage}
\end{tabular}
}
The axiom rule (\ref{simple_rules_ax}) shows that any basic category
reduces to itself~(identity).

\enumsentence{\label{simple_rules_ax}
	[Ax] \ $\lh${\X}:{\em t} $\rh$ \imp\ $\lh${\X}:{\em t} $\rh$}

\subsection{Qualia Structure in Categorial Grammar}

\paragraph{Application}

In (\ref{func-app-rev-left}) and (\ref{func-app-rev-right}) we present a
revised version of the rules for functional application. Note that the
difference between the rule for left application ({\X}/{\Y}) and right
application ({\Y}$\bs${\X}) is only in where the argument is found. Neither
the predicate-argument structure nor the qualia structure are
order-sensitive (see subsection~(\ref{CatSign}) at
page~\pageref{CatSign}).

\enumsentence{\label{func-app-rev-left}
\begin{tabular}[t]{ll}
[L/]
&
\begin{minipage}[t]{\textwidth}
	U, $\lh$ {\X}/{\Y} : ({\em f}, $[QS_f, QS_{a^{\prime}}]$)
	   $\rh$, T, V $\imp$ Z if
\\[1ex]
	T $\imp$ $\lh$ {\Y} : ({\em a}, $QS_a$) $\rh$ and
\\[1ex]
	U, $\lh$ {\X} : ({\em f(a)}, $[QS_f, QS]$)
           $\rh$, V $\imp$ Z\\
\end{minipage}
\end{tabular}
if {\Y} is basic then
	$QS$ = \{ $q \sqcup q^\prime \ | \
	q \epsilon QS_a \
	\& \ q^\prime \epsilon QS_{a^{\prime}} \
	\& \ q \sqcup q^\prime \neq \emptyset$ \};
	QS$_a$ otherwise
}

\enumsentence{\label{func-app-rev-right}
\begin{tabular}[t]{ll}
[L$\bs$]
&
\begin{minipage}[t]{\textwidth}
	U, T, $\lh$ {\Y}$\bs${\X} : ({\em f}, $[QS_f, QS_{a^{\prime}}]$)
	      $\rh$, V $\imp$ Z if
\\[1ex]
	T $\imp$ $\lh$ {\Y} : ({\em a}, $QS_a$) $\rh$ and
\\[1ex]
	U, $\lh$ {\X} : ({\em f(a)}, $[QS_f, QS]$)
           $\rh$, V $\imp$ Z\\
\end{minipage}
\end{tabular}
if {\Y} is basic then
	$QS$ = \{ $q \sqcup q^\prime \ | \
	q \epsilon QS_a \
	\& \ q^\prime \epsilon QS_{a^{\prime}} \
	\& \ q \sqcup q^\prime \neq \emptyset$ \};
	QS$_a$ otherwise
}

The functor usually has limited selection restrictions
($QS_{a^{\prime}}$) on its argument, whereas the argument itself tends
to have a more elaborate qualia structure ($QS_a$). A functor may for
instance require its argument to be of sort {\bf human\/}, whereas the
argument may have {\bf boy} as one value of its qualia structure, next
to a few other ones.

We take the qualia structure of the resulting type (the result of
applying the functor to the argument) to be the set of qualia roles of
the argument that unify with requirements of the functor. In other
words: that subset of the qualia structure of the argument that meets
some requirements of the functor.  From the set of qualia roles of the
argument it takes only those that unify with any of the requirements of
the functor.

The case where the unification of $QS_{a^{\prime}}$ and $QS_a$ is
empty---there is no value that unifies some qualia role from the
argument and some qualia role from the functor---indicates that no
interpretation can be found on the basis of the present information.
Such a restriction of the interpretation process to coercion from the
initial qualia structure, working only with stipulated values, implies
accepting that the process may fail at some point.

A sentence like (\ref{DowningStreet}), where an address should be
coerced to the function of the person who works there, the prime
minister, and from there to his or her spokesperson, will fail.

\enumsentence{\label{DowningStreet}
	Downing Street denied all knowledge today.}

That our theory does not capture all possible readings a sentence may
have, is in contrast with what we stated earlier, but we would need a
recursive application of coercion to avoid failing and derive values
also from the embedded qualia structures. Such an approach may be
computationally uncontrollable and we therefore pursue it no further
here\footnote{
	In (\ref{DowningStreet}) one might think of looking into the
	qualia structure of {\sl Downing Street}, finding the address
	of the Prime Minister; then looking into the qualia structure
	of {\sl Prime Minister} and finding an entry for {\sl
	spokesperson} there.}.

\paragraph{Abstraction}

The rules for abstraction are given in (\ref{ext_abstr_r}) and
(\ref{ext_abstr_l}). Since the resulting category {\X}/{\Y}
({\Y}$\bs${\X}, respectively) is hypothetical, the requirements
($QS_{a^{\prime}}$) for the embedded argument can not be lexically
given, but will result from the derivation. Similarly, the introduced
category {\Y} has only a hypothetical qualia structure $QS_a$, that must
be instantiated {\em on the fly}, i.e., through argument cancelation
elsewhere in the derivation.

\enumsentence{\label{ext_abstr_r}
\begin{tabular}[t]{ll}
[R/]
&
\begin{minipage}[t]{\textwidth}
	T $\imp$ $\lh$ {\X}/{\Y} : ({\em f}, $[QS_f, QS_{a^{\prime}}]$)
                  $\rh$	 if
\\[1ex]
\hspace*{10pt}
	T, $\lh$ {\Y} : ({\em a}, $QS_a$) $\rh$
		$\imp$
	$\lh$ {\X} : ({\em $\lambda$a.f}, $[QS_f, QS_{a^{\prime}}]$)
        $\rh$
\end{minipage}
\end{tabular}
}

\enumsentence{\label{ext_abstr_l}
\begin{tabular}[t]{ll}
[R$\bs$]
&
\begin{minipage}[t]{\textwidth}
	T $\imp$ $\lh$ {\Y}$\bs${\X} : ({\em f}, $[QS_f, QS_{a^{\prime}}]$)
        	 $\rh$	 if
\\[1ex]
\hspace*{10pt}
	$\lh$ {\Y} : ({\em a}, $QS_a$) $\rh$, T
		$\imp$
	$\lh$ {\X} : ({\em $\lambda$a.f}, $[QS_f, QS_{a^{\prime}}]$)
        $\rh$
\end{minipage}
\end{tabular}
}

\paragraph{Axiom}

The identity case, the axiom, is given in (\ref{ext_ax}). This now is
the place where the final type-check occurs. Note that this is {\em
not\/} where coercion takes place, coercion is typically a process
between functor and argument.

\enumsentence{\label{ext_ax}
\begin{tabular}[t]{ll}
[Ax]
&
\begin{minipage}{\textwidth}
	$\lh$ {\X} : ({\em t}, $QS_t$) $\rh$
		\imp\
	$\lh$ {\X} : ({\em t}, $QS_t$) $\rh$
\end{minipage}
\end{tabular}
}

\subsection{Examples}

\subsubsection{`begin a novel'}

In the following example we demonstrate how the transitive verb {\sl
begin} combines with the noun phrase {\sl a novel} to constitute the
verb phrase {\sl begin a novel}.

\enumsentence{\label{begin_a_novel}
\begin{minipage}[t]{\textwidth}
{\sl begin} :
 $\lh$ ({\NP}$\bs${\S})/ S/({\NP}$\bs${\S}) :
   ({\em $\lambda R \lambda x.R(\lambda y \exists e.begin(e,x,y))$}, \\[2ex]
\mbox{}\hspace*{5mm}
		\avm{QS$_{f}$:
			\avm{QS$_{f^\prime}$:\{x$_1$ \dots x$_n$\},\\
			     QS$_{a^\prime}$:\{\bf human\} },\\
		     QS$_{a}$:
			\avm{QS$_{f^{\prime\prime}}$:\{y$_1$ \dots y$_n$\},\\
			     QS$_{a^{\prime\prime}}$:
				\avm{QS$_{f^{\prime\prime\prime}}$:\{z$_1$ \dots z$_n$\},\\
				     QS$_{a^{\prime\prime\prime}}$:\{\bf event\} } }
		    }$)\rh$, \
\\[2ex]

{\sl a novel} :
 $\lh$ {\S}/({\NP}$\bs${\S}) : ({\em $\lambda P.\exists z.novel(z) \wedge
P(z)$}, \\[2ex]
\mbox{}\hspace*{5mm}	\avm{QS$_{f}$:\{x$_1$ \dots x$_n$\},\\
			     QS$_{a}$:
				\avm{QS$_{f^\prime}$:\{y$_1$ \dots y$_n$\},\\
				     QS$_{a^\prime}$:\{\bf artifact, read, write\}}}
) $\rh$ $\imp$
\\[2ex]
{\sl begin a novel} : $\lh$~{\NP}$\bs${\S} :
({\em $\lambda x.(\exists z.novel(z) \wedge (\exists e.begin(e,x,z)))$},\\[2ex]
\mbox{}\hspace*{5mm}
		\avm{QS$_{f}$:
			\avm{QS$_{f^\prime}$:\{x$_1$ \dots x$_n$\},\\
			     QS$_{a^\prime}$:\{\bf human\} },\\
		     QS$_{a}$:
			\avm{QS$_{f^{\prime\prime}}$:\{y$_1$ \dots y$_n$\},\\
			     QS$_{a^{\prime\prime}}$:
				\avm{QS$_{f^{\prime\prime\prime}}$:\{z$_1$ \dots z$_n$\},\\
				     QS$_{a^{\prime\prime\prime}}$:\{\bf read, write\} } }
		     }$)\rh$ \ \\[2ex]
where: {\bf read, write $\leq$ event}.
\end{minipage}
}
The functor, with category ({\NP}$\bs${\S})/{\NP}, applies to the
argument, with category {\NP}, to yield an {\NP}$\bs${\S}. The semantics
of the functor ({\em f\/}) applies to the semantics of the argument
({\em a\/}) and results in {\em f(a)\/}. The qualia structure of the
functor QS$_e$ remains unaffected, as well as the selection restrictions
of the subject argument (\{{\bf human}\}). The selection restrictions of
the object argument (\{{\bf event}\}) are intersected with the qualia
structure of the {\NP} (\{{\bf artifact, read, write}\}) which results
in \{{\bf read, write}\}, following {\bf read, write $\leq$ event}.

\subsubsection{`BMW announced \dots'}

This example demonstrates how a predicate combines with its subject
argument. The transitive verb `announced' combines with the noun phrase
`{\BMW}' to constitute the incomplete sentence `{\BMW} announced
$\dots$'\footnote{
	Another way to look at it would be to assume a {\VP}, and not a
	transitive verb, if one wishes to introduce bracketing. In that
	case the qualia structure of the object argument would have
	been specified. For our present purposes this is of no
	relevance.}

\enumsentence{\label{bmw_announced}
{\sl BMW} : $\lh$ {\NP} : ({\em $bmw$}, QS$_{a}$:\{{\bf company,
spokesperson,... }\}) $\rh$,
\\[2ex]
{\sl announced} : $\lh$ ({\NP}$\bs${\S})/{\NP} : ({\em $\exists e.\lambda
yx.announce(e,x,y)$}, \\[2ex]
\mbox{}\hspace*{5mm}
\avm{QS$_{f}$:
	\avm{QS$_{f^\prime}$:\{y$_1$ \dots y$_n$\},\\
		     QS$_{a^\prime}$:\{\bf human\}},\\
	     QS$_{a}$:\{\bf event\}}
\\[2ex]
{\sl BMW announced} : $\lh$ {\S}/{\NP} : ({\em $\exists e.\lambda
y.announce(e,bmw,y)$ },\\[2ex]
\mbox{}\hspace*{5mm}
\avm{QS$_{f}$:
	\avm{QS$_{f^\prime}$:\{y$_1$ \dots y$_n$\},\\
		     QS$_{a^\prime}$:\{\bf spokesperson\}},\\
	     QS$_{a}$:\{\bf event\}}
$\rh$
\\[2ex]
where: {\bf spokesperson $\leq$ human}.
}

The functor, with category ({\NP}$\bs${\S})/{\NP}, applies to the
argument, with category {\NP}, to yield an {\S}/{\NP}. The semantics of
the functor ({\em f\/}) applies to the semantics of the argument ({\em
a\/}) and results in {\em f(a)\/}. Again, the qualia structure of the
functor ($QS_e$) and the selection restrictions of the object argument
(\{{\bf event}\}) remain unaffected. The selection restrictions of the
subject argument (\{{\bf human}\}) are intersected with the qualia
structure of the {\NP} (\{{\bf company, spokesperson,\dots}\}), which
results in \{{\bf spokesperson}\}, following {\bf spokesperson $\leq$ human}.

%--------------------------6. Conclusion-------------------------------

\section{Conclusion}

The main criticism one may have against our work, is that it introduces
a new combinatorial explosion in Categorial Grammar, that already
suffers from the problem of spurious ambiguity\index{spurious
ambiguity} (\cite{hend:stud93}). A noun may have any finite number of
values in its qualia roles, and this may cause computational
problems. This is a valid objection. However, what we claim to derive is
all the possible readings a sentence could have. In other words, we show
how compositionality might work with functions that operate on
polysemous arguments. However, any research into the modelling of
constraints on interpretation that are set by our knowledge of the
(im)possibilities of the physical world is beyond the scope of this
paper.

Related to this, also the semantics~/~pragmatics interface remains a
problem to be solved. It is not yet clear to what extent pragmatic
inference is compositional or not. In this paper we tried to include
at least that part which can be dealt with compositionally\footnote{
	Some constraints on the interface of semantics and pragmatics may
	come from presuppositions. Bos, Buitelaar and Mineur
	(\shcite{bos:brid95}) study the parallel between qualia roles
	and presuppositions and takes coercion to operate on the
	presuppositions that lexical entries trigger. This does not
	effect the truth values for the entities that are denoted
	by these lexical items.}.

We did succeed however in increasing the semantic potential of
Categorial Grammar, while maintaining compositionality.  We have dealt
with polymorphism in lexical semantics. Instead of stipulating some
{\sf n} monomorphic types for one word, we adopted Pustejovsky's
approach to stipulate one polymorphic type which through coercion will
generate all {\sf n} monomorphic types. Such an approach explains in
part the creative use of language, which we find for instance in the
use of metonymy. On the other hand we acknowledged that Pustejovsky's
approach will not be helpful in cases of strict homonymy (bank/bank).

%--------------------------Acknowledgements----------------------------

\section{Acknowledgements}

This paper has benefitted from discussions with Hiyan Alshawi, Johan
Bos, Bob Carpenter, Ann Copestake, Kees van Deemter, David Milward,
Michael Moortgat, Gerald Penn, James Pustejovsky, Hans Uszkoreit, Marc
Verhagen, Kees Vermeulen, Leon Verschuur, our colleagues at Brandeis,
Saarbr\"ucken and Utrecht and from suggestions by an anonymous
reviewer. As always we are solely responsible for any remaining errors.
We would like to thank Kees van Deemter also for inviting us to write
this paper.

%----------------------------References--------------------------------

\end{document}